\documentclass{article}

\PassOptionsToPackage{numbers, compress}{natbib}
\usepackage[preprint]{neurips_2026}

\usepackage[utf8]{inputenc} 
\usepackage[T1]{fontenc}    
\usepackage{hyperref}       
\hypersetup{
    colorlinks=true,     
    linkcolor=blue,      
    citecolor=blue,      
    urlcolor=blue,       
    filecolor=blue       
}

\usepackage{url}            
\usepackage{booktabs}       
\usepackage{amsfonts}       
\usepackage{nicefrac}       
\usepackage{microtype}      
\usepackage{graphicx}
\usepackage{subcaption}
\usepackage{booktabs} 
\usepackage{multirow}
\usepackage{amsmath}
\usepackage{amssymb}
\usepackage{amsthm}
\usepackage{enumitem}
\usepackage{wrapfig}
\theoremstyle{plain}

\theoremstyle{definition}

\theoremstyle{remark}

\usepackage[textsize=tiny]{todonotes}

\usepackage[table]{xcolor}
\definecolor{qgray}{RGB}{240,240,240}
\definecolor{qblue}{RGB}{230,245,255}
\usepackage{tabularx}
\newcolumntype{Y}{>{\centering\arraybackslash}X} 

\title{LogitTrace: Detecting Benchmark Contamination via Layerwise Logit Trajectories}

\author{%
  Zirui He$^{1}$ \quad
  Haiyan Zhao$^{1}$ \quad
  Yingcong Li$^{1}$ \quad
  Ali Payani$^{2}$ \quad
  Mengnan Du$^{3}$\thanks{Corresponding author.} \\
  $^{1}$New Jersey Institute of Technology \\
  $^{2}$Cisco Research \\
  $^{3}$The Chinese University of Hong Kong, Shenzhen \\
  \texttt{\{zh296, hz54, yingcong.li\}@njit.edu} \\
  \texttt{apayani@cisco.com}, \quad \texttt{mengnandu@cuhk.edu.cn}
}

\begin{document}

\maketitle

\begin{abstract}
Large language models (LLMs) are commonly evaluated on challenging benchmarks such as \emph{AIME} and \emph{Math500}, where benchmark contamination can make memorized solutions appear as genuine reasoning. Existing detection methods largely rely on surface overlap, completion behavior, or final-output likelihood, and often degrade when inputs are simply rephrased. In this paper, we propose \textbf{LogitTrace} (Layerwise Logit Trajectories), a framework for analyzing memorization-like decision dynamics through intermediate logit trajectories. Instead of judging memorization only from the final answer, LogitTrace examines how model preferences emerge and stabilize across layers. We find that contaminated examples tend to show earlier commitment, while clean examples exhibit more gradual evidence accumulation. These trajectory signals allow a lightweight classifier to separate contaminated and clean examples across multiple models and input variants. Controlled LoRA injection experiments further show that repeated exposure to target samples induces similar trajectory patterns. Overall, our results suggest that LogitTrace provides evidence beyond surface overlap and final-output confidence, offering a useful lens for studying memorization-like behavior in LLMs.
\end{abstract}

\section{Introduction}

Large language models (LLMs) have achieved remarkable performance across reasoning, dialogue, and code generation tasks~\citep{brown2020language,openai2023gpt4,touvron2023llama}. However, their success has raised growing concerns about \emph{data contamination}, where benchmark test data (or their transformed variants) appear in the training corpus. Such contamination can lead models to reproduce memorized solutions while appearing to reason, compromising claims of true generalization and raising privacy and intellectual property risks~\citep{carlini2021extracting,zhao2024measuring,nasr2025scalable}. Contamination has been documented in widely used datasets such as MMLU and TruthfulQA~\citep{deng2024contamination}. It can yield deceptively strong outputs: models may still solve compressed or even invalid problems, reflecting recall rather than reasoning. Alternatively,~\citet{wu2025reasoning} demonstrate that reinforcement learning gains on math benchmarks can vanish once contaminated items are removed, underscoring the urgent need for systematic contamination detection.

Existing detection methods primarily rely on completion-based recall tests \citep{carlini2023quantifying,nasr2025scalable} and distributional probes such as perplexity and output agreement \citep{li2023perplexity,dong2024generalization,shi2024detecting,zhang2024divergence,zhang2024pacost}. Other approaches include adversarial compression \citep{schwarzschild2024memorization} and partial probing \citep{zhao2024measuring,ye2024datacali}. While these methods provide valuable evidence of contamination, they often depend on surface forms, completion behavior, or final-output statistics.  As a result, their robustness can degrade when test problems are rephrased, paraphrased, translated, or structurally altered \citep{deng2024contamination,yao2024crosslingual}. For example, \citet{yang2023rethinking} report that adding paraphrased MMLU items into training raises Llama 2 accuracy from 45-55\% to nearly 89\%, almost matching GPT-4. Such findings reveal that contamination can persist even without verbatim overlap, motivating more robust representation-level detection methods.

\begin{figure*}[t]
    \centering
    \includegraphics[width=\textwidth]{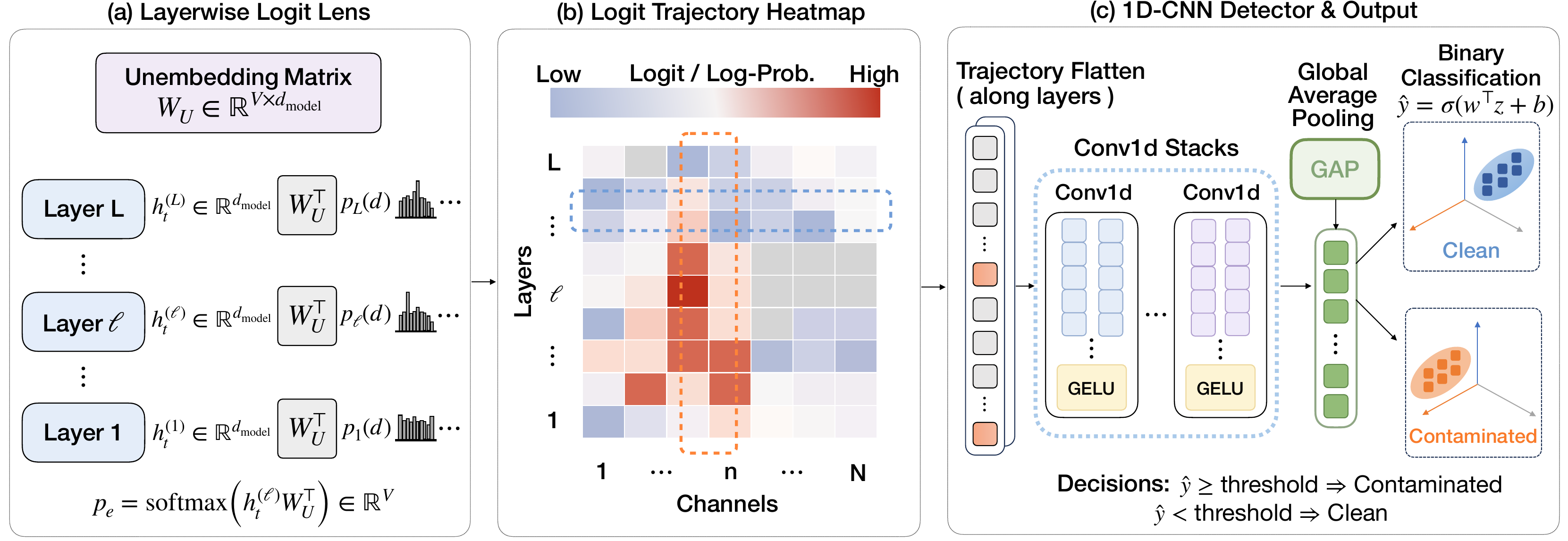}
    \caption{
Overview of LogitTrace.
(a) Intermediate hidden states are projected through the unembedding matrix to obtain layerwise digit probabilities.
(b) For each sample, these trajectories form a layer-by-channel feature matrix, with rows denoting layers and columns denoting trajectory features.
(c) The resulting matrix is fed as a multichannel sequence into a lightweight 1D-CNN detector, followed by global average pooling and a binary classifier for clean/contaminated prediction.
}
    \label{fig:logittrace_overview}
\end{figure*}

To address these challenges, we introduce \textbf{LogitTrace} (Layerwise Logit Trajectories), a detection framework that identifies contamination-associated signals from layerwise logit dynamics. The key idea is that memorization may affect not only what the model answers, but also how the answer forms across layers. LogitTrace therefore tracks intermediate logit trajectories and tests whether their temporal structure separates contaminated from clean examples. We find a consistent pattern: contaminated examples tend to commit earlier, while clean examples accumulate evidence more gradually. A lightweight classifier built on these trajectories outperforms completion- and likelihood-based baselines under rephrasing, translation, and perturbation. Controlled LoRA injection experiments further show that repeated exposure to target samples strengthens similar trajectory signatures. Our contributions are threefold:
\begin{itemize}[leftmargin=*]\setlength\itemsep{-0.1em}
    \item We introduce LogitTrace, a detector that identifies contamination-associated decision dynamics from layerwise logit trajectories.
    \item We show that early commitment patterns distinguish contaminated from clean examples across multiple models and input variants.
    \item We use controlled LoRA injection to test whether repeated exposure induces the same type of trajectory signal observed in contaminated examples.
\end{itemize}

\section{Preliminary}
\label{sec:memorization-definitions}

\subsection{A Model-Level View of Memorization}

Let $\Sigma$ be the vocabulary and $\Sigma^\star$ the set of all finite sequences. A causal language model $M_\theta$ defines a conditional distribution $p_\theta(y \mid x)$ over continuations $y \in \Sigma^\star$ given a prompt $x \in \Sigma^\star$. We define the sequence-level log-likelihood as
\begin{equation}
\label{eq:seq-loglik}
s_\theta(y \mid x) \triangleq \sum_{t=1}^{|y|} \log p_\theta\!\big(y_t \,\big|\, x, y_{<t}\big).
\end{equation}

Conceptually, a sequence $y^\star$ is memorized by $M_\theta$ if there exists a context $x$ under which the model assigns $y^\star$ high likelihood and prefers it over plausible alternatives. Formally, for thresholds $\tau>0$ and $\kappa \ge 0$,
\begin{equation}
\label{eq:memorization}
\frac{1}{|y^\star|}\, s_\theta(y^\star \mid x) \ge \tau, \quad
s_\theta(y^\star \mid x) - \max_{y \neq y^\star} s_\theta(y \mid x) \ge \kappa.
\end{equation}
The first condition captures high likelihood, while the second imposes a margin against alternative continuations.

This view treats memorization as a model-level behavior rather than a directly observed dataset property. In modern LLMs, we usually cannot verify whether a benchmark item actually appeared in training data. We therefore use operational contamination labels and controlled injection experiments to study observable signatures associated with memorization-like behavior.

\subsection{Explicit vs. Implicit Contamination}
\label{sec:contamination-definitions}

We study contamination at the data level, where benchmark items or their variants may appear in training data and affect evaluation. Following the model-level view above, such exposure can manifest during inference as memorization-like behavior. Since the exact training corpus is usually unavailable, we distinguish contamination regimes by the observable relationship between an evaluated benchmark item and a potentially exposed counterpart.

\paragraph{Explicit contamination.}
We refer to a benchmark item as \emph{explicitly contaminated} when the evaluated instance preserves substantial surface overlap with a possibly exposed training instance. In this regime, detectors based on lexical similarity, completion behavior, or likelihood heuristics can often provide useful evidence~\citep{deng2024contamination,nasr2025scalable}.

\paragraph{Implicit contamination.}
We refer to a benchmark item as \emph{implicitly contaminated} when contamination effects persist even after the evaluated instance is transformed to reduce surface overlap, such as through rephrasing, translation, or structural perturbation. Recent findings show that surface-level signals are fragile in this setting: rephrased or translated benchmark items can still induce contamination effects even without verbatim overlap~\citep{yang2023rethinking,dong2024generalization}. This motivates detection strategies that capture subtler memorization dynamics beyond surface-level cues.

\subsection{Representation Probing}
\label{sec:rep-eng}

Intermediate representations can reveal predictive information that is not visible from the final answer alone. The logit lens projects hidden states through the language modeling head to obtain layerwise vocabulary distributions~\citep{nostalgebraist2020logitlens}, while the tuned lens improves this view with learned affine probes~\citep{belrose2025elicitinglatentpredictionstransformers}. Related inversion and representation-decoding work further shows that next-token distributions and internal representations can preserve recoverable information about the input~\citep{morris2024language,zhao2026rep2textdecodingtextsingle}. Motivated by these findings, LogitTrace tracks answer-relevant digit probabilities across layers and tests whether contaminated samples exhibit distinct layerwise dynamics from clean samples.

\section{Methodology}\label{sec:method}

\subsection{Problem Formulation}\label{sec:prob-form}

We consider an LLM $M$ that takes a textual problem $x$ as input and produces an answer $a$. Our objective is to detect whether an example $(x,a)$ should be classified as \emph{contaminated} or \emph{clean} under an operational contamination setting. Formally, we construct a binary classification problem from a dataset $\{(x_i,a_i,y_i)\}_{i=1}^N$, where $y_i \in \{0,1\}$ denotes the operational contamination label: $y_i=1$ for contaminated examples and $y_i=0$ for clean examples.

This formulation allows us to design a representation-based detector $f_\theta$ that takes intermediate activations from $M$ as input and predicts a contamination score.

\begin{figure*}[t]
    
    \centering 

    \includegraphics[width=0.9\textwidth]{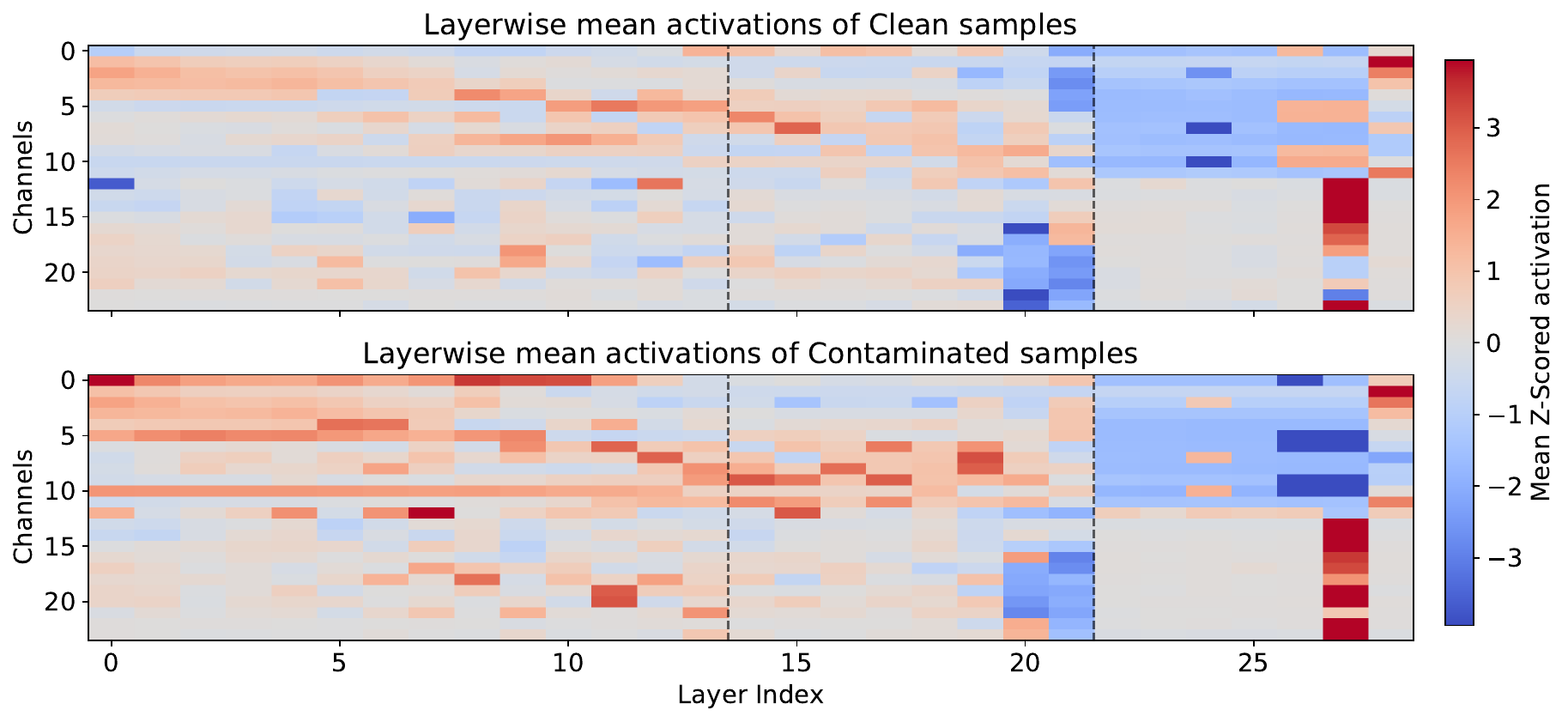}
    
    \caption{Class-mean activation heatmaps.
Average channel activations across layers for clean vs. contaminated samples.
Red regions denote stronger activation and blue denotes weaker activation.}
    
    \label{fig:heatmap}
    
\end{figure*}

\subsection{Representation Extraction}\label{sec:rep-extra}

Building on the logit-lens view introduced in Section~\ref{sec:rep-eng}, we extract digit-level trajectories from the hidden states of $M$. Let $h_\ell \in \mathbb{R}^d$ denote the residual stream activation at layer $\ell$. Applying the language modeling head $W_\text{LM}$ yields logits and a predictive distribution over the vocabulary:
\begin{equation}\label{eq:logit-prob}
    z_\ell = h_\ell W_\text{LM}, \quad p_\ell = \mathrm{softmax}(z_\ell).
\end{equation}

Since we focus on numerical reasoning tasks, we restrict this distribution to digit tokens. Let $\mathcal{V}_d$ denote the set of token IDs corresponding to digit $d \in \{0,\dots,9\}$. The digit-only probability at layer $\ell$ is defined as:
\begin{equation}
    p_\ell(d) = \frac{\sum_{t \in \mathcal{V}_d} p_\ell(t)}{\sum_{j=0}^9 \sum_{t \in \mathcal{V}_j} p_\ell(t)}.
\end{equation}

From this digit-level distribution, we compute two scalar statistics: digit entropy $H_\ell$ and maximum digit probability $m_\ell$:
\begin{equation}\label{eq:entro-max}
    H_\ell = - \sum_{d=0}^9 p_\ell(d) \log p_\ell(d), \quad m_\ell = \max_{d} p_\ell(d).
\end{equation}
Together, these quantities describe how concentrated the model's digit preferences are at each layer, providing a compact view of layerwise commitment during answer formation.

\subsection{Feature Construction}\label{sec:fea-con}
\label{chapter:features}

For each sample $(x,a)$, we identify the first generation position corresponding to the answer and extract trajectories across the first $T$ layers. At each layer $\ell$, we concatenate digit probabilities, entropy, maximum digit probability, and their first-order differences:
\begin{equation}\label{eq:feature}
    X = \big[ p_\ell(0..9),\; H_\ell,\; m_\ell,\; \Delta p_\ell,\; \Delta H_\ell,\; \Delta m_\ell \big]_{\ell=1}^T,
\end{equation}
where $\Delta$ denotes first-order temporal differences, e.g., $\Delta H_\ell = H_\ell - H_{\ell-1}$. This results in a multichannel sequence representation $X \in \mathbb{R}^{C \times T}$, with $C=24$ channels: 12 raw features and 12 difference features.

\subsection{CNN-based Detector}\label{sec:cnn-disc}

We train a lightweight 1D convolutional neural network with hidden dimension $d=128$ to classify whether a trajectory $X$ is labeled as contaminated or clean. 
The network consists of three convolutional layers with ReLU activation and a global average pooling layer. 
The classifier produces logits
\begin{equation}\label{eq:cnn}
    z = \text{CNN}(X) \in \mathbb{R}^d, \, o = W z + b \in \mathbb{R}^2, \, \hat y = \text{softmax}(o).
\end{equation}
The training objective is the standard cross-entropy loss, where $y_{i,c}$ denotes the one-hot operational label:
\begin{equation}
    \mathcal{L}(\theta) = - \sum_{i=1}^N \sum_{c=1}^2 y_{i,c} \log \hat y_{i,c}.
\end{equation}
This detector tests whether the temporal structure of layerwise logit trajectories contains information that separates contaminated and clean examples beyond hand-crafted rules or single-layer statistics.

\begin{table*}[t]
  \caption{
    Performance comparison of different methods across four text distributions. All numerical results are reported to one decimal place. Cont. means contaminated. 
  }
  \label{tab1}
  \footnotesize
  \begin{tabularx}{\textwidth}{l *{8}{Y}} 
    \toprule
    & \multicolumn{2}{c}{Original} & \multicolumn{2}{c}{Rephrased} & \multicolumn{2}{c}{Translated} & \multicolumn{2}{c}{Perturbed} \\
    \cmidrule(lr){2-3} \cmidrule(lr){4-5} \cmidrule(lr){6-7} \cmidrule(lr){8-9}
    & Cont. $\uparrow$ & Clean $\downarrow$ & Cont. $\uparrow$ & Clean $\downarrow$ & Cont. $\uparrow$ & Clean $\downarrow$ & Cont. $\uparrow$ & Clean $\downarrow$ \\
    \midrule
    
    \rowcolor{qgray}
    \multicolumn{9}{l}{\normalsize\textit{\textbf{Qwen-2.5-7B}}} \\
    Completion-based    & 100.0 & 11.6 & 16.7 & 3.7 & 12.2 & 2.6 & 15.8 & 3.0 \\
    TS-Guessing         & 92.7 & 14.5 & 21.3 & 15.0 & 3.6 & 0.0 & 17.7 & 14.6 \\
    Perplexity          & 87.6 & 1.1 & 29.2 & 0.4 & 10.8 & 0.7 & 19.0 & 0.0 \\
    Output Distribution & 64.7 & 28.0 & 62.6 & 45.3 & 26.4 & 16.5 & 58.9 & 39.3 \\
    \rowcolor{qblue}
    LogitTrace (Ours)   & 95.6 & 5.6 & 63.1 & 2.2 & 20.0 & 1.5 & 60.1 & 1.5  \\
    \midrule 
    
    \rowcolor{qgray}
    \multicolumn{9}{l}{\normalsize\textit{\textbf{Qwen-2.5-Math-7B}}} \\
    Completion-based    & 100.0 & 8.7 & 18.3 & 3.7 & 13.6 & 2.6 & 15.4 & 1.9 \\
    TS-Guessing         & 96.4 & 11.4 & 19.5 & 13.9 & 1.5 & 0.0 & 17.1 & 12.4 \\
    Perplexity          & 96.1 & 3.8 & 38.4 & 3.7 & 30.9 & 3.0 & 26.7 & 3.0  \\
    Output Distribution & 71.3 & 19.3 & 50.5 & 20.9 & 17.8 & 11.2 & 50.5 & 26.6 \\
    \rowcolor{qblue}
    LogitTrace (Ours)   & 89.4 & 5.6 & 56.2 & 2.6 & 64.2 & 9.4 & 51.7 & 1.9  \\
    \midrule
    
    \rowcolor{qgray}
    \multicolumn{9}{l}{\normalsize\textit{\textbf{Llama-3.1-8B}}} \\
    Completion-based    & 100.0 & 14.4 & 4.3 & 6.3 & 3.2 & 3.4 & 2.9 & 4.8  \\
    TS-Guessing         & 98.2 & 17.3 & 29.1 & 0.3 & 0.3 & 0.2 & 0.3 & 0.2  \\
    Perplexity          & 93.7 & 11.2 & 36.1 & 10.6 & 31.5 & 8.7 & 24.9 & 8.1 \\
    Output Distribution & 60.0 & 18.0 & 59.8 & 36.4 & 39.3 & 20.2 & 62.8 & 34.8 \\
    \rowcolor{qblue}
    LogitTrace (Ours)   & 95.7 & 12.8 & 63.0 & 4.1 & 46.4 & 4.5 & 61.0 & 4.5  \\
    \midrule
    
    \rowcolor{qgray}
    \multicolumn{9}{l}{\normalsize\textit{\textbf{Qwen-3-8B}}} \\
    Completion-based    & 100.0 & 14.2 & 19.4 & 2.6 & 13.8 & 2.6 & 18.8 & 1.9 \\
    TS-Guessing         & 91.3 & 21.7 & 20.1 & 19.5 & 3.8 & 1.9 & 19.0 & 17.6 \\
    Perplexity          & 70.0 & 39.7 & 63.7 & 42.3 & 84.4 & 64.8 & 57.5 & 36.7 \\
    Output Distribution & 78.0 & 33.0 & 83.8 & 33.3 & 33.1 & 19.1 & 64.4 & 32.9 \\
    \rowcolor{qblue}
    LogitTrace (Ours)   & 87.5 & 14.8 & 93.1 & 13.1 & 78.1 & 10.1 & 64.4 & 13.9 \\
    \bottomrule
  \end{tabularx}
\end{table*}

\section{Experiments}

In this section, we evaluate the performance of \textbf{LogitTrace} on memorization detection tasks to address the following research questions:
\begin{itemize}[leftmargin=*]\setlength\itemsep{-0.1em}
    \item \textbf{RQ1:} Do layerwise logit trajectories provide reliable signals for separating contaminated and clean examples, and are these signals robust to input variations?

    \item \textbf{RQ2:} Does controlled exposure through LoRA fine-tuning increase LogitTrace's contamination scores and induce similar trajectory patterns?

    \item \textbf{RQ3:} How do these     contamination-associated signals depend on layer depth and detector architecture, and do they transfer to external benchmarks?
\end{itemize}

\subsection{Experimental Setup}
\label{section4.1}
\paragraph{Models and Datasets.}
We evaluate four open-source LLMs including {Qwen-2.5-7B}, {Qwen-2.5-Math-7B}, {Qwen-3-8B}, and Llama-3.1-8B. Contaminated samples are constructed from \emph{AIME 1983--2024} and \emph{Math500} using a masked-completion procedure: the model is given partial problem text, and samples are retained when the generated completion exactly reconstructs the target text. Clean samples are constructed from \emph{AIME 2025}, \emph{Minerva Math}, and \emph{LiveMathBench}, which were released after these models. To evaluate robustness, we additionally introduce three variants: \textit{Rephrased} (paraphrased wording), \textit{Perturbed} (surface-level edits), and \textit{Translated} (French translation), by prompting GPT-4o-mini. We report the prompt details in Appendix~\ref{appendix:b}.

\paragraph{Baselines.}
We compare against four prior approaches. 
(i) \textit{Completion-based}~\citep{carlini2023quantifying}: a sample is flagged as contaminated if the model's greedy completion exactly reconstructs the target text. 
(ii) \textit{TS-Guessing}~\citep{deng2024contamination}: contamination is detected if the model can recover the target text from partial-information prompts or symbolic transformations. 
These two are completion-driven baselines. 
(iii) \textit{Perplexity}~\citep{li2023perplexity}: computes the log-likelihood of the target text under the model; unusually low perplexity is taken as evidence of memorization. 
(iv) \textit{Output Distribution}~\citep{dong2024generalization}: measures the similarity between greedy decoding and multiple stochastic completions; higher similarity indicates stronger memorization-like behavior. 
These two are distribution-based baselines.

\paragraph{Metrics.}
We report detection metrics for the main contamination experiments. 
(i) \textit{Youden's J index}: $J=\mathrm{TPR}-\mathrm{FPR}$, used to select the detector threshold by balancing contaminated detection and clean false positives.
(ii) \textit{Contamination Detection Rate}: the fraction of contaminated examples correctly flagged as contaminated at the chosen threshold.
(iii) \textit{False Positive Rate}: the fraction of clean examples incorrectly flagged as contaminated.

For the LoRA injection experiment, we additionally report reconstruction metrics to measure whether controlled exposure improves verbatim recovery of the target text. Specifically, we use a masked-completion setup in which the model observes a partial problem statement and is asked to complete the original text.
(iv) \textit{Exact Match (EM / pass@1)}: the fraction of first-generation completions that exactly match the target text.
(v) \textit{Rouge-L}: the longest common subsequence overlap between the generated completion and the target text, measuring partial reconstruction.

\begin{table}[t]
\centering
\caption{Detailed performance of LogitTrace across evaluation settings, reporting AUROC, PR-AUC, Precision, Recall, and F1. }
\label{tab:logittrace_supp_metrics}
\setlength{\tabcolsep}{6pt}
\renewcommand{\arraystretch}{1.12}
\begin{tabular*}{\linewidth}{@{\extracolsep{\fill}} llccccc @{}}
\toprule
Model & Setting & AUROC & PR-AUC & Precision & Recall & F1 \\
\midrule
\multirow{4}{*}{Qwen-2.5-7B}
& Original   & 0.982 & 0.985 & 0.973 & 0.956 & 0.964 \\
& Rephrased  & 0.944 & 0.972 & 0.983 & 0.631 & 0.769 \\
& Translated & 0.837 & 0.912 & 0.966 & 0.200 & 0.332 \\
& Perturbed  & 0.948 & 0.974 & 0.988 & 0.601 & 0.748 \\
\midrule
\multirow{4}{*}{Qwen-2.5-Math-7B}
& Original   & 0.962 & 0.979 & 0.975 & 0.894 & 0.933 \\
& Rephrased  & 0.918 & 0.961 & 0.980 & 0.520 & 0.679 \\
& Translated & 0.900 & 0.953 & 0.944 & 0.642 & 0.764 \\
& Perturbed  & 0.922 & 0.962 & 0.985 & 0.517 & 0.678 \\
\midrule
\multirow{4}{*}{Llama-3.1-8B}
& Original   & 0.958 & 0.925 & 0.855 & 0.929 & 0.890 \\
& Rephrased  & 0.927 & 0.890 & 0.902 & 0.630 & 0.742 \\
& Translated & 0.865 & 0.804 & 0.862 & 0.464 & 0.603 \\
& Perturbed  & 0.925 & 0.885 & 0.899 & 0.610 & 0.727 \\
\midrule
\multirow{4}{*}{Qwen-3-8B}
& Original   & 0.905 & 0.804 & 0.778 & 0.875 & 0.824 \\
& Rephrased  & 0.960 & 0.929 & 0.810 & 0.931 & 0.866 \\
& Translated & 0.927 & 0.881 & 0.822 & 0.781 & 0.801 \\
& Perturbed  & 0.949 & 0.910 & 0.796 & 0.900 & 0.845 \\
\bottomrule
\end{tabular*}
\end{table}

\subsection{Main Results}
\label{sec:mainresults}
\paragraph{Quantitative Analysis.}
We evaluate all methods on four text distributions: original, rephrased, translated, and perturbed. Table~\ref{tab1} shows clear differences across detection strategies. Completion-based approaches (Completion-based and TS-Guessing) perform well on original problems, but degrade sharply once the inputs are rephrased or perturbed, with contaminated detection rates often dropping below 20\%. Distribution-based methods (Perplexity and Output Distribution) are somewhat less tied to exact completions, but remain unstable: Perplexity can produce high false positives on clean samples, while Output Distribution varies substantially across models and input variants. In contrast, LogitTrace generally provides a stronger trade-off between detecting contaminated examples and avoiding false positives on clean ones. For example, on rephrased Qwen-2.5-7B data, completion-based detection reaches only 16.7\%, whereas LogitTrace reaches 63.1\%; on perturbed Llama-3.1-8B data, Output Distribution mislabels over 30\% of clean cases, while LogitTrace limits this to 4.5\%.

Table~\ref{tab:logittrace_supp_metrics} further reports threshold-independent and classification metrics for LogitTrace. Across most models and input variants, LogitTrace achieves high AUROC and PR-AUC, suggesting that the trajectory scores provide useful ranking signals beyond a single threshold. Precision is also consistently high, indicating that examples flagged as contaminated are rarely clean false positives. The main degradation appears in recall and F1 under translated inputs for some models, especially Qwen-2.5-7B, suggesting that cross-lingual transformations remain more challenging than rephrasing or perturbation. Overall, these results indicate that layerwise logit trajectories provide robust contamination-associated signals.

\begin{table*}[t]
  \centering
  \caption{
Detection and reconstruction results under LoRA fine-tuning with varying rank $r$. 
Contamination Detection Rate (CDR) denotes the fraction of target samples flagged as contaminated by the fixed detector.
EM, Rouge-L, and Contamination Detection Rate are reported as percentages. 
}
  \label{tab:simple_results}
  \small
  \begin{tabular*}{\textwidth}{@{\extracolsep{\fill}}l l c c c c l}
    \toprule
    &\hspace{1.5em}Model & Trainable-parameters & EM (pass@1) & RougeL & CDR&\\
    \midrule
    &\hspace{1.5em}Pre-trained Model   & 0    & 17.8 & 35.4 & 2.2 &\\
    \midrule
    &\multicolumn{5}{l}{\normalsize\textit{\hspace{1.4em}\textbf{Fine-tuned Model}}} &\\
    &\hspace{1.5em}$r=8$ (32 epochs)     & 20M   & 27.2 & 53.0 & 14.9 &\\
    &\hspace{1.5em}$r=16$ (32 epochs)    & 40M   & 41.8 & 65.5 & 19.0 &\\
    &\hspace{1.5em}$r=32$ (32 epochs)    & 80M   & 52.2 & 73.6 & 29.7 &\\
    &\hspace{1.5em}$r=64$ (32 epochs)    & 170M  & 58.9 & 77.6 & 38.0 &\\
    &\hspace{1.5em}$r=128$ (32 epochs)   & 340M  & 69.8 & 81.6 & 39.9 &\\
    &\hspace{1.5em}$r=256$ (32 epochs)   & 670M  & 75.9 & 88.1 & 44.3 &\\
    &\hspace{1.5em}$r=512$ (32 epochs)   & 1.3B  & 78.5 & 89.1 & 45.1 &\\
    \bottomrule
  \end{tabular*}
\end{table*}

\begin{figure*}[t]
    
    \centering 

    \includegraphics[width=0.99\textwidth]{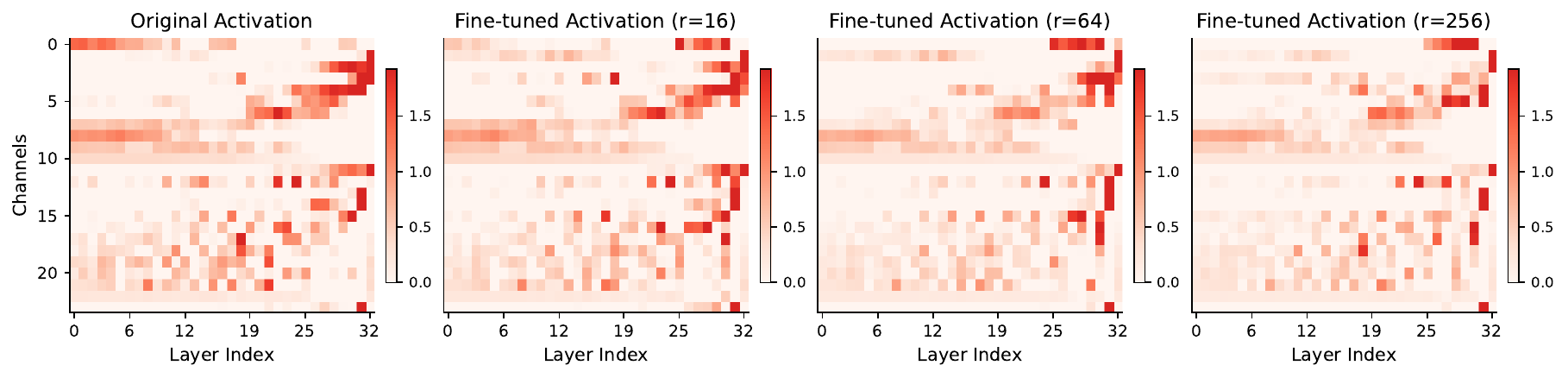}
    
    \caption{Layer-wise activations for one sample across LoRA ranks, with layers on the x-axis, 24 feature channels on the y-axis, and color showing normalized values.}
    
    \label{fig:heatmap_lora}
    
\end{figure*}

\paragraph{Representation Analysis.}
We further examine class-mean activation heatmaps to understand how LogitTrace separates contaminated from clean examples. In Figure~\ref{fig:heatmap}, the x-axis denotes model layers, the y-axis denotes the 24 trajectory feature channels, and color indicates the class-mean z-scored activation for each channel-layer pair. Compared with clean examples, contaminated examples show more pronounced high-activation regions in the early and middle layers. This pattern supports the layerwise-commitment hypothesis: contaminated examples tend to form answer-related preferences earlier in the forward pass.

\subsection{LoRA Injection Evaluation}

To test whether LogitTrace is sensitive to trajectory changes induced by controlled fine-tuning exposure, we conduct \emph{LoRA injection experiments}. 
The detector $f_\theta$ is trained once and kept fixed throughout this experiment. 
We then fine-tune the base model $M_{\text{base}}$ on the same target samples using LoRA with different ranks $r$, obtaining $M_{\text{lora}}$. 
A larger LoRA rank introduces more trainable adaptation parameters, allowing the fine-tuned model to fit the target samples more strongly. 
After fine-tuning, we apply the same fixed detector to the same samples under each LoRA model. 
If the detector score increases with $r$, this indicates that LogitTrace is sensitive to trajectory changes induced when the model has been trained on those samples.

As shown in Table~\ref{tab:simple_results}, under $M_{\text{base}}$, these samples are almost always classified as clean, with a Contamination Detection Rate of only 2.2\%. 
After LoRA fine-tuning, however, a larger fraction of samples are identified as contaminated, with the rate increasing to 44.3\% at $r=256$ and 45.1\% at $r=512$. 
This upward trend at small-to-moderate $r$ is also aligned with reconstruction performance: both EM (pass@1) and Rouge-L steadily improve as $r$ grows, suggesting that fine-tuning makes the model more capable of reconstructing the target text. 
Overall, the parallel increase in reconstruction performance and Contamination Detection Rate suggests that controlled exposure through LoRA fine-tuning strengthens the contamination-associated trajectory signals detected by LogitTrace.

In Figure~\ref{fig:heatmap_lora}, we present a controlled case study tracing a single evaluation sample across LoRA ranks. 
With the detector $f_\theta$ and threshold fixed, the predicted contamination probability increases systematically from 0.20 for the base model to 0.62 ($r{=}16$), 0.92 ($r{=}64$), and 0.98 ($r{=}256$). 
Since the detector, prompts, and test sample remain unchanged, this monotonic increase provides controlled causal evidence that LoRA exposure amplifies the trajectory signal used by the detector. 
The heatmaps illustrate how this signal emerges: in the base model, the feature channels remain relatively diffuse in the early layers without a clear preference. 
After LoRA fine-tuning, however, early-layer activations become more concentrated, and this concentration becomes more pronounced as $r$ grows.

\begin{figure*}[t]
\centering
\begin{minipage}[t]{0.47\textwidth}
  \centering
  \includegraphics[height=4.1cm]{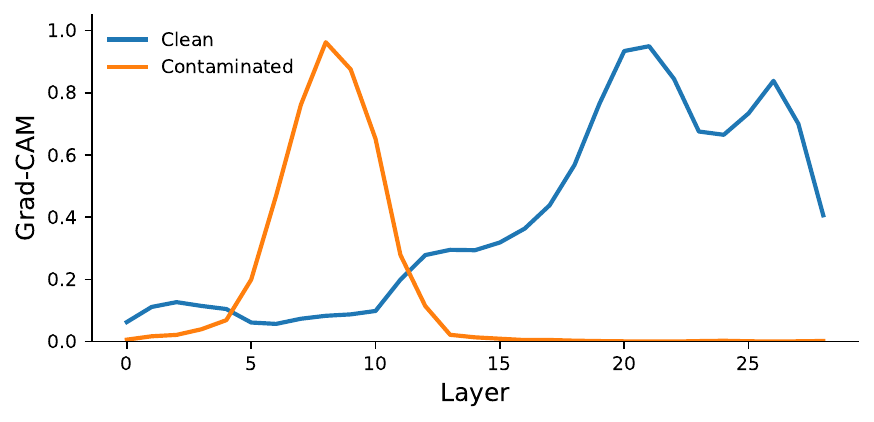}
  \caption{Class-mean Grad-CAM visualization comparing clean vs. contaminated samples.}
  \label{fig:gradcam}
\end{minipage}
\hspace{0.04\textwidth}
\begin{minipage}[t]{0.45\textwidth}
  \centering
  \includegraphics[height=3.65cm]{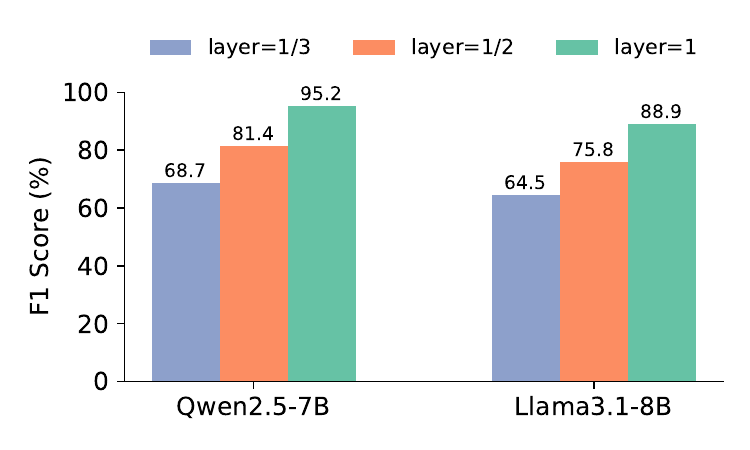}
  \caption{F1 scores across different layer depths.}
  \label{fig:layer_ablation}
\end{minipage}
\end{figure*}

\subsection{Ablation Study}

\paragraph{Layer Selection.}
In the main experiments, detectors use activations from the full model depth. To assess where the relevant signal appears, we train additional detectors using only the first one-third or first half of the layers. As shown in Figure~\ref{fig:layer_ablation}, the first one-third of layers already provides meaningful performance, with F1 scores of 68.7\% on Qwen-2.5-7B and 64.5\% on Llama-3.1-8B. Using the first half improves the scores to 81.4\% and 75.8\%, while using the full depth gives the best results, reaching 95.2\% and 88.9\%, respectively. These results suggest that contamination-associated signals emerge early, but aggregating evidence across depth yields the most reliable detector. Grad-CAM visualizations in Figure~\ref{fig:gradcam} support this trend: contaminated samples show sharper importance peaks in shallow layers, while clean samples have smoother importance distributions.

\paragraph{Detector Selection.}
To assess sensitivity to the detector architecture, we compare three lightweight detectors operating on the same multichannel sequence representation $X$ under the same data split: our 1D-CNN detector, a two-layer MLP that averages features across depth, and a small two-layer Transformer encoder.
As shown in Table~\ref{tab:detector_ablation}, the 1D-CNN provides the best overall trade-off, achieving strong AUROC/PR-AUC (0.960/0.977) and the lowest False Positive Rate (5.56\%).
The MLP performs substantially worse, indicating that collapsing depth-wise trajectories discards critical information.
The Transformer achieves comparable AUROC/PR-AUC and higher Contaminated Detection Rate, but incurs a substantially higher False Positive Rate (16.67\%), suggesting a less favorable clean-side trade-off.
Architectural and training details are deferred to Appendix~\ref{appendix:d}.

\paragraph{External Benchmark Generalization.}
Prior work has reported substantial contamination risks in MMLU, making it a useful external benchmark for testing whether LogitTrace captures contamination-associated structure beyond our constructed evaluation sets. 
We apply the detector trained on \textit{Qwen-2.5-Math-7B} to the \emph{mathematics subset of MMLU} in a zero-shot setting, without retraining or changing the detector threshold. 
Since item-level contamination labels are unavailable, we use the detector outputs for grouping: samples predicted as contaminated are referred to as \emph{seen-like}, while those predicted as clean are referred to as \emph{not-seen}.

\begin{figure*}[t]
  \centering
  \includegraphics[width=0.33\textwidth]{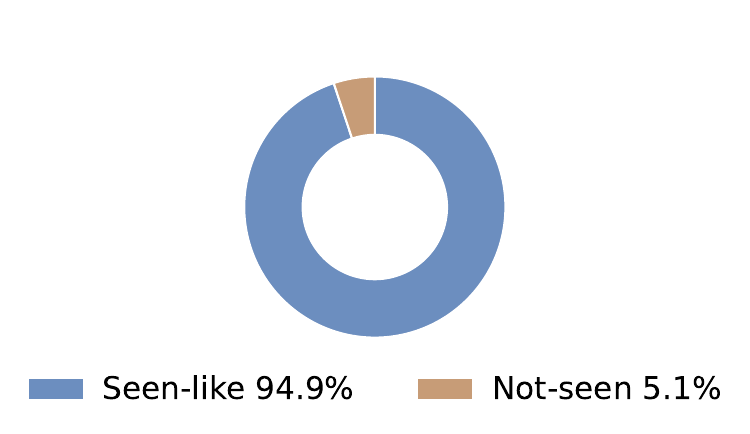}%
  \includegraphics[width=0.33\textwidth]{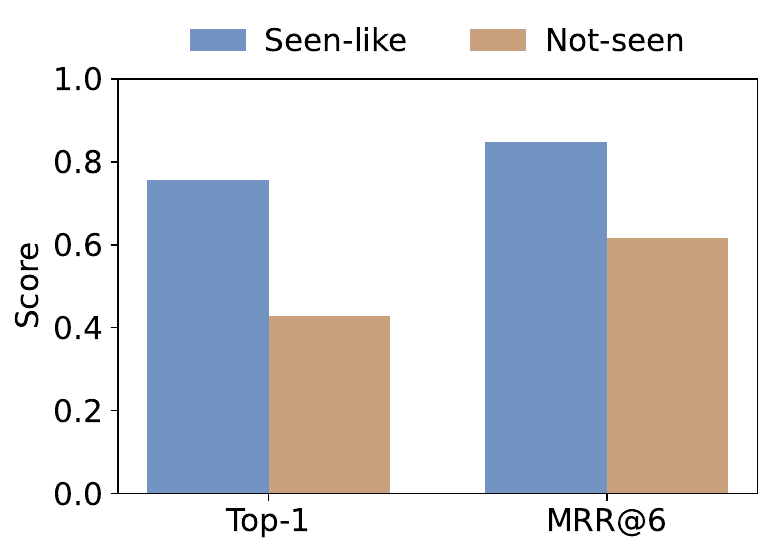}%
  \includegraphics[width=0.3\textwidth]{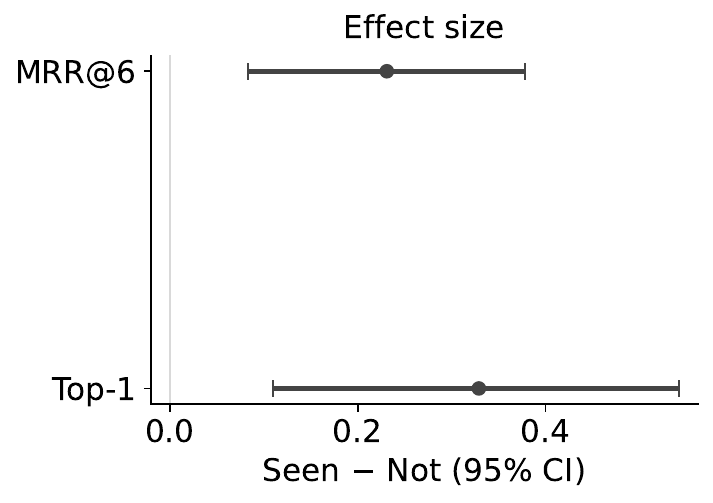}
  \caption{
    Dataset composition and group-level performance comparison. 
\textbf{Left:} fraction of samples classified as seen-like (94.9\%) versus not-seen (5.1\%). 
\textbf{Middle:} Top-1 accuracy and MRR@6 for the two groups. 
\textbf{Right:} effect sizes (Seen–Not) with 95\% confidence intervals. }
  \label{fig:mmlu_analysis}
\end{figure*}

As shown in Figure~\ref{fig:mmlu_analysis}, LogitTrace labels 94.9\% of the problems as \emph{seen-like} and 5.1\% as \emph{not-seen}, which is consistent with prior reports that MMLU contains substantial contamination risk. 
More importantly, this split corresponds to a clear behavioral difference: \textit{Qwen-2.5-Math-7B} achieves substantially higher performance on the \emph{seen-like} group, with Top-1 accuracy higher by 0.33 and MRR@6 higher by 0.23. 
The confidence intervals indicate that these gaps are statistically reliable. 
These results suggest that LogitTrace transfers to an external benchmark and identifies a seen-like axis that correlates strongly with model performance, supporting its use as a tool for analyzing benchmark familiarity and contamination-associated behavior.

\begin{table}[t]
\centering
\small
\caption{Detector architecture ablation using identical trajectory features. Cont. Detection Rate is the fraction of contaminated examples correctly flagged, while False Positive Rate is the fraction of clean examples incorrectly flagged.}
\label{tab:detector_ablation}
\setlength{\tabcolsep}{5pt}
\renewcommand{\arraystretch}{1.08}
\begin{tabular}{lccccc}
\toprule
Detector & Params (M) & AUROC & PR-AUC & Cont. Detection Rate (\%) & False Positive Rate (\%) \\
\midrule
CNN         & 0.079 & 0.960 & 0.977 & 84.09 & 5.56  \\
MLP         & 0.003 & 0.504 & 0.632 & 87.88 & 87.04 \\
Transformer & 0.069 & 0.954 & 0.974 & 95.45 & 16.67 \\
\bottomrule
\end{tabular}
\end{table}

\section{Related Work}

\paragraph{Contamination.}
Benchmark results are often inflated when training and test overlap. Leakage has been documented in MMLU and other benchmarks \citep{zhou2023don,deng2024contamination}, while rephrased and translated variants can raise accuracy without verbatim overlap \citep{yang2023rethinking,yao2024crosslingual}. Evaluation-side defenses attempt to mitigate these risks: CLEAN-EVAL and inference-time filtering sanitize benchmarks \citep{zhu2023cleaneval,zhu2024itd}, DCQ provides automated auditing \citep{golchin2025dcq}, and task-level contamination undermines few-shot claims \citep{li2024taskcontamination}. 

\paragraph{Detection and Mechanisms.}
Detection protocols formalize memorization as extractable or discoverable and extend to compression-based probing with shorter triggers \citep{carlini2023quantifying,nasr2025scalable,schwarzschild2024memorization}, which capture verbatim recall yet still underestimate hidden forms under longer or rephrased inputs. Statistical calibration methods broaden coverage: minimum-probability detectors \citep{shi2024detecting} and divergence-based scoring \citep{zhang2024divergence} flag anomalous likelihoods, paired confidence tests improve statistical reliability \citep{zhang2024pacost}, and black-box calibration extends applicability to proprietary models \citep{ye2024datacali}. Interventions can offer stronger causal evidence. For example, DICE operationalizes contamination by fine-tuning paired contaminated and uncontaminated models and then detecting in-distribution leakage from their internal representations \citep{tu2024dice} and temporal audits tracing when benchmarks leaked into training corpora \citep{golchin2024time}. Beyond detection, boundary studies interrogate whether high performance reflects reasoning or memorization, through idiom understanding \citep{kim2025idiom}, knowledge updating \citep{li2025kup}, generalization failures \citep{dong2024generalization}, and overestimation in machine translation \citep{kocyigit2025overestimation}. Together, these studies suggest that, despite advances in coverage and robustness, existing methods remain largely tied to surface cues or distributional anomalies. This limitation motivates representation-level analysis that can more directly expose the underlying dynamics of memorization.

\section{Conclusion}

In this paper, we introduce LogitTrace, a representation-based framework for detecting benchmark contamination in LLMs. By analyzing layerwise logit trajectories, LogitTrace provides reliable signals that remain stable under rephrasing and perturbation. Our comprehensive experiments, including controlled LoRA injection, not only confirm the robustness of this approach but also provide causal evidence that these trajectory patterns reflect model exposure rather than surface-level artifacts. Overall, LogitTrace offers a practical lens for studying memorization-like behavior and reassessing benchmark contamination in LLMs.

\bibliographystyle{plainnat}
\bibliography{custom}

\clearpage

\appendix
\section{Full CoT Outputs}
\label{appendix:a}

These four examples illustrate the potential risk that memorization may masquerade as generalization. In Table ~\ref{appendix-table:5}, the model answers the original problem correctly, which is expected. In Table ~\ref{appendix-table:6}, after rephrasing, the model still provides the correct solution, appearing to generalize. In Table ~\ref{appendix-table:7}, however, even when the input is reduced to a minimal set of keywords, the model produces the correct answer together with a full chain-of-thought explanation. This behavior is unlikely to arise from genuine reasoning and instead suggests direct recall of memorized content. In Table ~\ref{appendix-table:8}, the model attempts to generate an answer even for an invalid input, further underscoring its reliance on pattern matching rather than  reasoning. These cases reveal that model outputs may reflect hidden memorization signals disguised as generalization, which poses potential risks in downstream use.

\vspace{40pt}

\begin{table}[h]
\centering
\small

\caption{Case study 1: Original baseline question.}
\label{appendix-table:5}
\begin{tabular}{p{0.95\linewidth}}
\toprule
\textbf{Case Study 1: Original Question (Baseline)} \\
\midrule
\textbf{Question:} \\
A regular pentagon is rotated counterclockwise about its center. What is the minimum number of degrees it must be rotated until it coincides with its original position? \\
\midrule
\textbf{Model Output:} \\
72 degrees.

To determine the minimum number of degrees a regular pentagon must be rotated counterclockwise about its center until it coincides with its original position, we need to consider the symmetry of the pentagon. A regular pentagon has rotational symmetry, meaning it looks the same after being rotated by certain angles around its center.

The order of rotational symmetry for a regular pentagon is 5. This means there are 5 different positions where the pentagon looks exactly the same when rotated. The angle of rotation for each symmetry position is given by dividing 360 degrees (a full rotation) by the order of symmetry.

So, we calculate:
\[
\frac{360^\circ}{5} = 72^\circ
\]

Therefore, the minimum number of degrees the pentagon must be rotated counterclockwise until it coincides with its original position is \(\boxed{72}\). \\
\bottomrule
\end{tabular}
\end{table}

\vspace{40pt}
\begin{table}[!htbp]
\centering
\caption{Case study 2: Hexagon input, model outputs correct result.}
\label{appendix-table:6}
\begin{tabular}{p{0.95\linewidth}}
\toprule
\textbf{Case Study 2: Core Number Changed (Hexagon)} \\
\midrule
\textbf{Question:} \\
A regular hexagon is rotated counterclockwise about its center. What is the minimum number of degrees it must be rotated until it coincides with its original position? \\
\midrule
\textbf{Model Output:} \\
60 degrees.

To determine the minimum number of degrees a regular hexagon must be rotated counterclockwise about its center until it coincides with its original position, we need to consider the symmetry of the hexagon. A regular hexagon has rotational symmetry, meaning it looks the same after being rotated by certain angles around its center.

A regular hexagon has 6 equal sides and 6 equal angles. The sum of the interior angles of a hexagon is given by the formula \((n-2) \times 180^\circ\), where \(n\) is the number of sides. For a hexagon (\(n=6\)), the sum of the interior angles is:
\[
(6-2) \times 180^\circ = 4 \times 180^\circ = 720^\circ
\]
However, we are interested in the angles of rotation that will map the hexagon onto itself. Since the hexagon has 6 equal sides, it can be divided into 6 equal parts. Therefore, the angle of rotation that will map the hexagon onto itself is:
\[
\frac{360^\circ}{6} = 60^\circ
\]
This means that if the hexagon is rotated by 60 degrees counterclockwise, it will coincide with its original position. any multiple of 60 degrees (i.e., 120 degrees, 180 degrees, 240 degrees, 300 degrees, 360 degrees, etc.) will also map the hexagon onto itself, but the minimum positive angle is 60 degrees.

Thus, the minimum number of degrees the hexagon must be rotated until it coincides with its original position is \(\boxed{60}\). \\
\bottomrule
\end{tabular}
\end{table}

\begin{table}[t]
\centering
\caption{Case study 3: Keyword-based input, model still outputs the correct reasoning.}
\label{appendix-table:7}
\begin{tabular}{p{0.95\linewidth}}
\toprule
\textbf{Case Study 3: Keyword Summary Input} \\
\midrule
\textbf{Question:} \\
minimum rotation degrees, regular pentagon, symmetry \\
\midrule
\textbf{Model Output:} \\
\(72^\circ\)

The minimum rotation degrees that will map a regular pentagon onto itself is \(72^\circ\). This is because a regular pentagon has five sides, and a full rotation of \(360^\circ\) divided by five gives \(72^\circ\). This means that rotating the pentagon by \(72^\circ\) will bring it back to a position where it looks exactly the same as it did before the rotation.

To understand this better, let's consider the symmetry of a regular pentagon. A regular pentagon has five lines of symmetry, each passing through a vertex and the midpoint of the opposite side. When the pentagon is rotated by \(72^\circ\), each vertex moves to the position of the next vertex in a clockwise or counterclockwise direction. This rotation aligns the pentagon with itself, leaving it unchanged.

Therefore, the minimum rotation degrees that will map a regular pentagon onto itself is \(\boxed{72^\circ}\). \\
\bottomrule
\end{tabular}
\end{table}

\begin{table}[!htbp]
\centering
\caption{Case study 4: Input with non-integer polygon sides.}
\label{appendix-table:8}
\begin{tabular}{p{0.95\linewidth}}
\toprule
\textbf{Case Study 4: Input Floating Number (4.5)} \\
\midrule
\textbf{Question:} \\
A regular 4.5-sided polygon is rotated counterclockwise about its center. What is the minimum number of degrees it must be rotated until it coincides with its original position? \\
\midrule
\textbf{Model Output:} \\
80 degrees.

To determine the minimum number of degrees a regular 4.5-sided polygon must be rotated until it coincides with its original position, we start by understanding the properties of regular polygons and their rotational symmetry.

A regular polygon with \( n \) sides can be rotated by \( \frac{360^\circ}{n} \) degrees to coincide with its original position. This is because a full rotation of \( 360^\circ \) degrees will bring the polygon back to its starting position, and dividing this by the number of sides gives the smallest angle of rotation that will result in the polygon coinciding with itself.

In this problem, the polygon has 4.5 sides. However, a polygon must have an integer number of sides. Therefore, we need to interpret the problem as referring to a polygon with a fractional number of sides in a theoretical or conceptual sense, and we can use the same formula to find the angle of rotation.

The formula for the angle of rotation is:
\[
\text{Angle of rotation} = \frac{360^\circ}{4.5}
\]

To simplify this, we can convert 4.5 to a fraction:
\[
4.5 = \frac{9}{2}
\]

Now, substitute this into the formula:
\[
\text{Angle of rotation} = \frac{360^\circ}{\frac{9}{2}} = 360^\circ \times \frac{2}{9} = \frac{360^\circ \times 2}{9} = \frac{720^\circ}{9} = 80^\circ
\]

Thus, the minimum number of degrees the polygon must be rotated until it coincides with its original position is:
\[
\boxed{80}
\] \\
\bottomrule
\end{tabular}
\end{table}

\clearpage

\section{Rephrase dataset prompt}
\label{appendix:b}
\subsection{Problem Transformation Setup}

We applied controlled prompting to generate different variants of math problems for our experiments. All generations were obtained using the \texttt{GPT-4o-mini} model (via the \texttt{chat.completions} API). To ensure variability while preserving mathematical correctness, we set the sampling temperature to 0.7.

We designed three transformation tasks:
\begin{itemize}
    \item \textbf{Rephrasing}: synonym replacement or structure change without altering semantics.
    \item \textbf{Perturbation}: minor linguistic noise while keeping numbers/LaTeX fixed.
    \item \textbf{Translation (French)}: translate only the surrounding text, keeping numbers/LaTeX unchanged.
\end{itemize}

The exact prompts used for each task are shown in Table~\ref{tab:prompts}.

\begin{table}[h]
\centering
\caption{Prompt templates used to generate rephrased, perturbed, and translated math problems. 
All generations used GPT-4o-mini with temperature $=0.7$.}
\begin{tabular}{p{0.18\linewidth} p{0.77\linewidth}}
\toprule
\textbf{Task} & \textbf{Prompt Template} \\
\midrule
Rephrase & 
Rephrase the following math problem in English. Requirements: \newline
1. Keep ALL numbers and LaTeX expressions exactly the same. \newline
2. Do not change the mathematical meaning. \newline
3. Output ONLY the rephrased problem, no explanations. \newline
Problem: \{q\} \\
\midrule
Perturb & 
Rewrite the following math problem with minor linguistic perturbations. You may change word order, replace synonyms, or alter sentence structure. Requirements: \newline
1. Keep ALL numbers and LaTeX expressions exactly unchanged. \newline
2. Preserve the exact mathematical meaning. \newline
3. Output ONLY the perturbed problem. \newline
Problem: \{q\} \\
\midrule
Translate (French) & 
Translate the following math problem into French. Requirements: \newline
1. Keep ALL numbers and LaTeX expressions unchanged. \newline
2. Translate only the surrounding text. \newline
3. Output ONLY the translated problem, no explanations. \newline
Problem: \{q\} \\
\bottomrule
\end{tabular}
\label{tab:prompts}
\end{table}

\clearpage
\section{More Activation Trajectories Across Models}
\label{appendix:c}

\begin{figure}[!h]
    \centering 
    \includegraphics[width=0.75\textwidth]{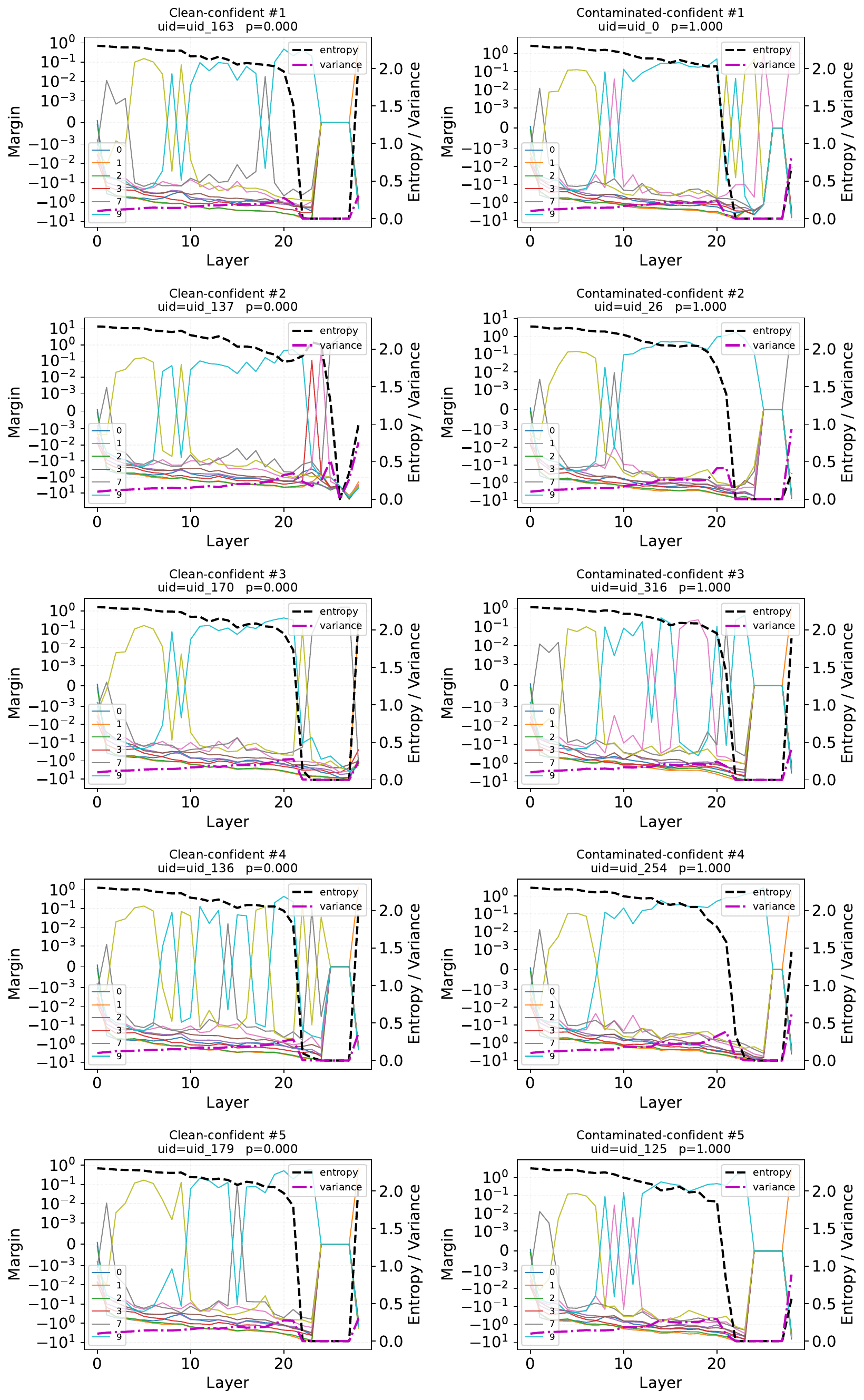}
    \caption{Activation trajectories of the 10 most confidently classified samples by the detector trained on Qwen-2.5-7B. Clean samples are shown on the left, while contaminated samples are shown on the right.
}   
\end{figure}

\begin{figure}[t]
    \centering 
    \includegraphics[width=0.75\textwidth]{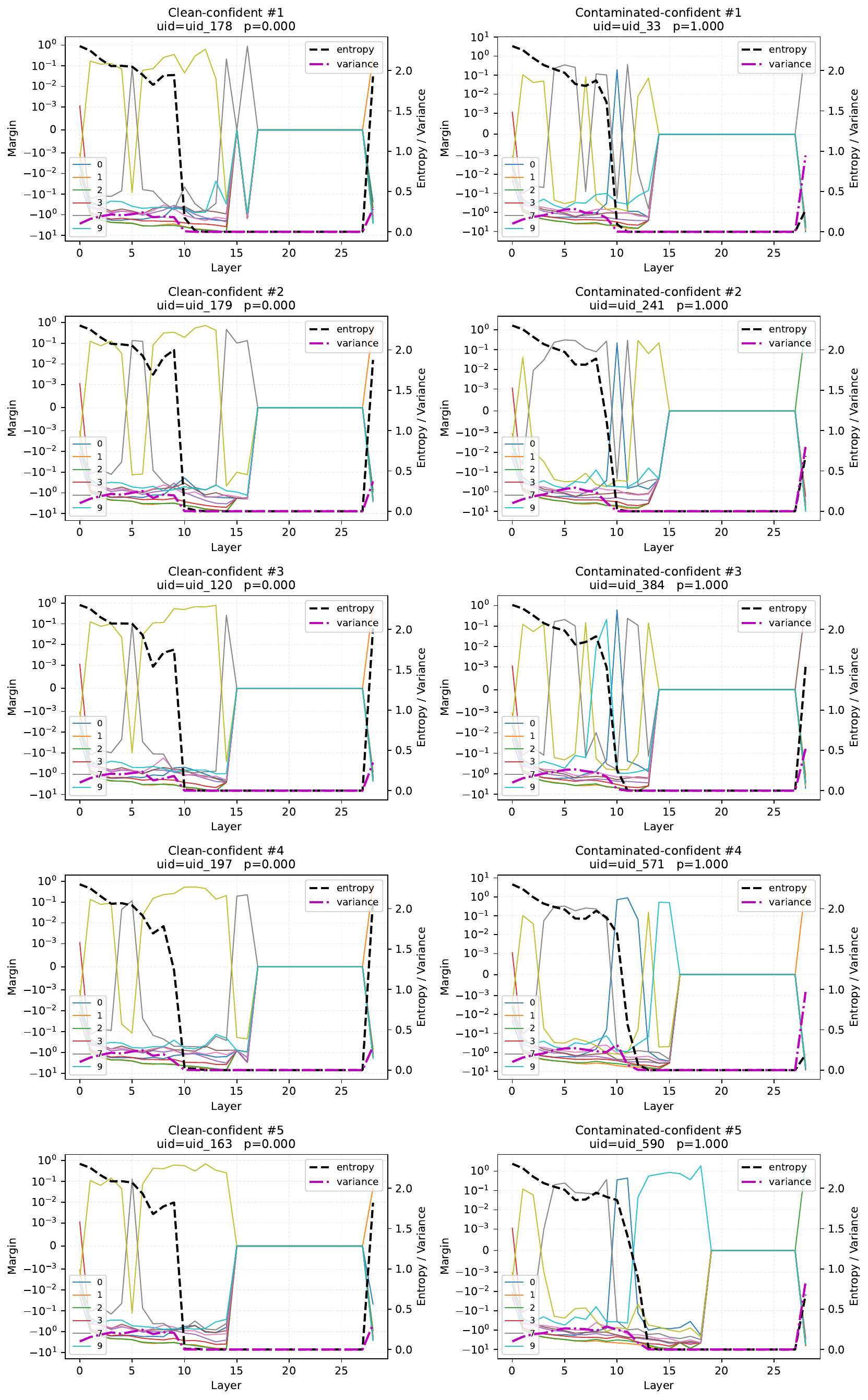}
    \caption{Activation trajectories of the 10 most confidently classified samples by the detector trained on Qwen-2.5-Math-7B. Clean samples are shown on the left, while contaminated samples are shown on the right.
}
\end{figure}

\begin{figure}[h]
    \centering 
    \includegraphics[width=0.75\textwidth]{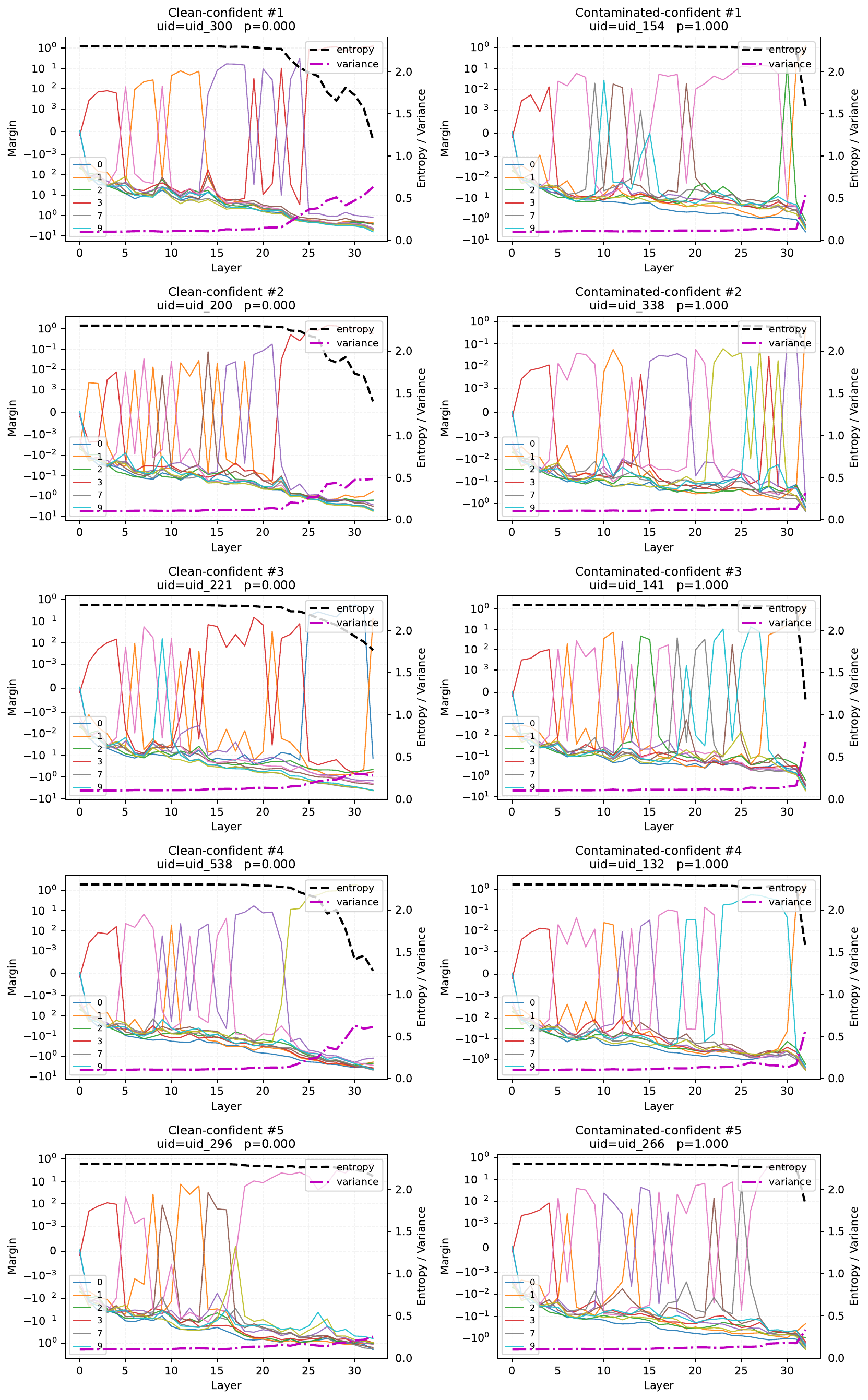}   
    \caption{Activation trajectories of the 10 most confidently classified samples by the detector trained on Llama-3.1-8B. Clean samples are shown on the left, while contaminated samples are shown on the right. 
}    
\end{figure}

\begin{figure}[h]
    
    \centering 

    \includegraphics[width=0.75\textwidth]{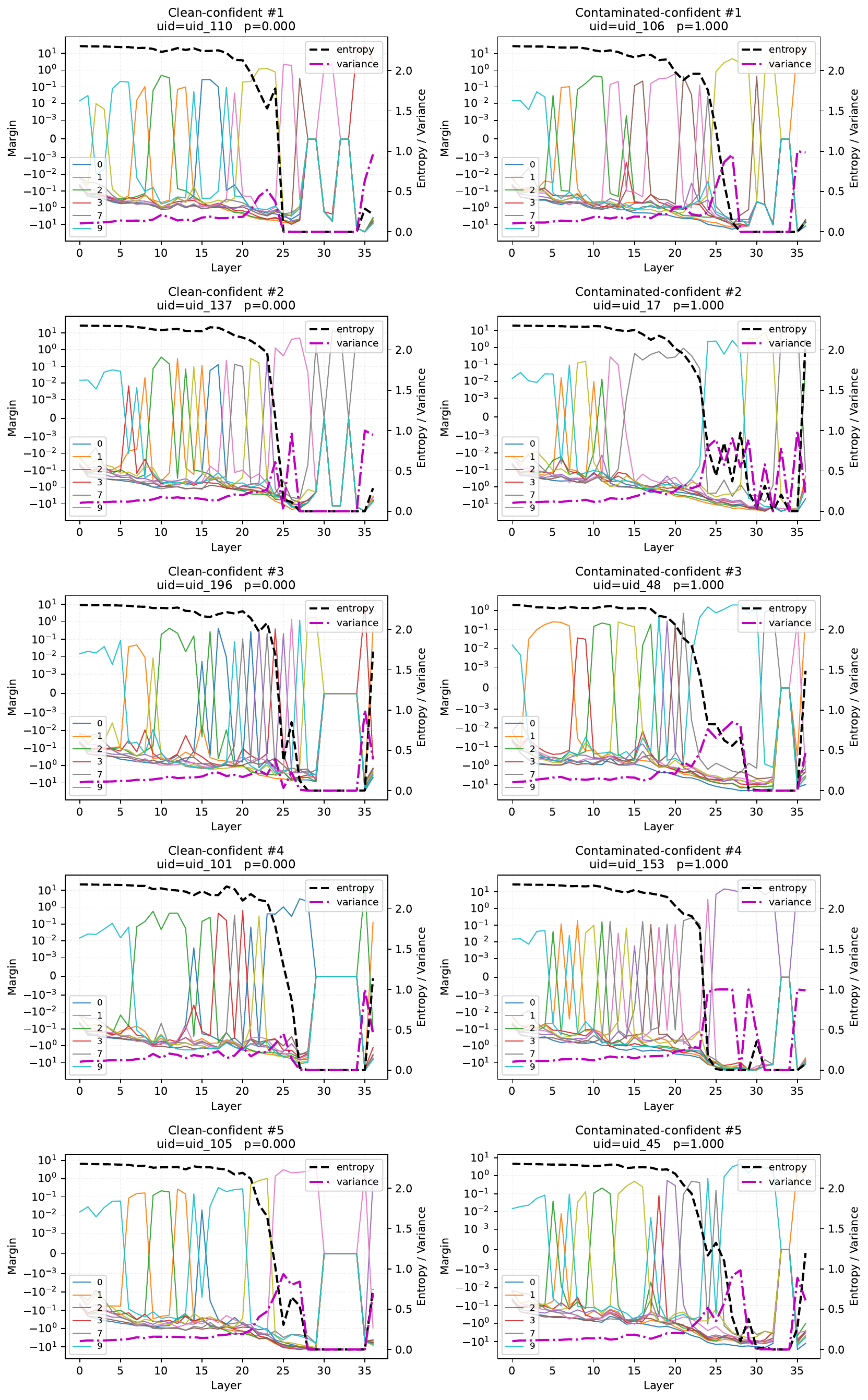}
    
    \caption{Activation trajectories of the 10 most confidently classified samples by the detector trained on Qwen-3-8B. Clean samples are shown on the left, while contaminated samples are shown on the right.
}

\end{figure}

\clearpage
\section{Implementation Details}
\label{appendix:d} 

\subsection{Detector Training}
\paragraph{Feature construction.}
For each problem–answer pair, we tokenize the prompt \texttt{``Problem: <q>\textbackslash nAnswer:''} with the model’s native tokenizer and run the model with \texttt{output\_hidden\_states=True}. At every layer $l$, we compute (i) the normalized probability mass over digit tokens $p_{l}(d)$ for $d\in\{0,\dots,9\}$ (token sets collected from the tokenizer vocabulary), (ii) the entropy $H_l$ of $p_l$, and (iii) the maximum probability $\max_{d} p_l(d)$. We also take first-order differences along depth for each of these streams. Stacking gives a $C\times T$ trajectory with $C=24$ channels:
10 digit channels $p_{l}(d)$, plus $H_l$ and $\max p_l$, and their 12 first-order differences. We z-score normalize \emph{per sample, per channel} (NaNs $\to$ 0). For sequences with different $T$, we truncate all samples to the minimal $T$ across the batch to align time axes.

\paragraph{Detector architecture.}
We use a lightweight 1D CNN (\texttt{TinyTSConv}) with three \texttt{Conv1d} layers (kernel size $3$, channels $C\!\to\!64\!\to\!128\!\to\!128$) and ReLU activations, followed by global average pooling and a 2-way linear head. Inputs have shape $[N,C,T]$ with $C=24$.

\paragraph{Training protocol.}
We perform a stratified split with validation ratio $0.2$.
Optimization uses AdamW with learning rate $2\times 10^{-3}$, batch size $32$, and cross-entropy loss.
To address class imbalance, we apply inverse-frequency class weights
$\displaystyle w_k=\frac{n_{\text{pos}}+n_{\text{neg}}}{2\,\max(n_k,1)}$.
We train for $15$ epochs and select the checkpoint that maximizes AUROC on the validation set.
All random seeds are fixed to $42$ (Python/NumPy/PyTorch). Training runs on the same device as the backbone model (CPU/GPU), with features stored as \texttt{float32}.

\paragraph{Detector ablation.}  Regarding the choice of a CNN-based detector, we conducted an additional ablation study comparing the CNN with a two-layer MLP and a lightweight 2-layer Transformer encoder ($d_{\text{model}}=64$, $n_{\text{head}}=4$), all trained on the same representation-trajectory features and under the same train/validation split.

The MLP aggregates the $[C, T]$ feature map by averaging over the temporal dimension to produce a $C$-dimensional vector, followed by two linear layers with a ReLU activation in between.

The Transformer detector begins by projecting each layer-wise feature vector from $C$ to a 64-dimensional embedding. This sequence of $T$ embeddings is then fed into a small encoder consisting of two 4-head self-attention layers with a 128-dimensional feed-forward block and the standard GELU activation used in PyTorch's implementation. After encoding, the model averages over the temporal dimension and applies a final linear classifier. Both alternatives have comparable parameter counts to the CNN (approximately 0.003M for the MLP and 0.07M for the Transformer versus 0.079M for the CNN).

\subsection{Lora Fine-tuning Experiment}
\paragraph{LoRA fine-tuning.}
We fine-tune \textit{Llama-3.1-8B} using LoRA adapters applied to attention and MLP projection layers 
(\texttt{q\_proj}, \texttt{k\_proj}, \texttt{v\_proj}, \texttt{o\_proj}, \texttt{gate\_proj}, \texttt{up\_proj}, \texttt{down\_proj}). 
Training is performed in 4-bit NF4 quantization with double quantization, and all experiments vary the LoRA rank $r$ while setting $\alpha=2r$ and dropout $0.05$. 
Problems are concatenated and segmented into blocks of 1024 tokens, where the model is trained in the standard causal language modeling objective with cross-entropy loss. 
We use AdamW with learning rate $1\!\times\!10^{-4}$, cosine decay, warmup ratio $0.03$, batch size $2$ per device with gradient accumulation $8$, and train for $32$ epochs. 
Gradient checkpointing and TF32 are enabled. 
All random seeds are fixed to $0$ for reproducibility.

\section{More Related Work}
\label{appendix:e}
\paragraph{Copyright.} A practical data contamination scenario is copyright. DE-COP identifies copyrighted content in pretraining corpora \citep{duarte2024decop}, audits reveal explicit copyright violations in model outputs \citep{karamolegkou2023copyright}, partial probing quantifies exposure risk \citep{zhao2024measuring}, and further tests examine whether models comply with copyright notices or reproduce licensed code \citep{xu2024respectcopyright,yu2023codeipprompt}. Such findings underscore that contamination is not only a technical artifact but also a legal and ethical risk: reproducing protected material can mislead benchmark evaluations and at the same time threaten compliance with intellectual property.

\section{Limitations}

LogitTrace uses operational contamination labels because the true pretraining corpus of each model is unavailable, LogitTrace relies on operational contamination labels. Still, the consistency of its trajectory patterns across models, input variants, and controlled exposure experiments suggests that the detector captures meaningful contamination-associated signals. Its predictions should therefore be interpreted as diagnostic evidence rather than proof that a sample appeared in training data. The current experiments focus on numerical reasoning and digit-level trajectories, so extensions to open-ended or non-numeric tasks may require new task-specific token sets and feature definitions. In addition, LogitTrace requires access to intermediate model representations and layerwise logits, which limits its direct applicability to closed-source models that expose only final outputs.


\end{document}